\documentclass[sigplan,screen]{acmart}

\renewcommand\footnotetextcopyrightpermission[1]{} 
\setcopyright{none}

\usepackage[utf8]{inputenc}
\usepackage{xspace}
\usepackage{caption}
\usepackage{pifont}
\usepackage{flushend}
\usepackage{subfig}
  
\begin{document}









  \title{\huge \textsc{MixNN}\xspace: Protection of Federated Learning Against Inference Attacks by Mixing Neural Network Layers}



\author{Antoine Boutet}
\affiliation{
	\institution{Univ Lyon, INSA Lyon, Inria, CITI}
}
\email{antoine.boutet@insa-lyon.fr}

\author{Thomas Lebrun}
\affiliation{%
  \institution{Univ Lyon, INSA Lyon, Inria, CITI}
}
\email{thomas.lebrun@inria.fr}

\author{Jan Aalmoes}
\affiliation{%
  \institution{Univ Lyon, INSA Lyon, Inria, CITI}
}
\email{jan.aalmoes@inria.fr}

\author{Adrien Baud}
\affiliation{%
  \institution{Univ Lyon, INSA Lyon, Inria, CITI}
}
\email{adrien.baud@inria.fr}

  \begin{abstract}
{
  Machine Learning (ML) has emerged as a core technology to provide learning models to perform complex tasks.
Boosted by Machine Learning as a Service (MLaaS), the number of applications relying on ML capabilities is ever increasing.
However, ML models are the source of different privacy violations through passive or active attacks from different entities.
In this paper, we present \name a proxy-based privacy-preserving system for federated learning to protect the privacy of participants against a curious or malicious aggregation server trying to infer sensitive attributes.
\name receives the model updates from participants and mixes layers between participants before sending the mixed updates to the aggregation server.
This mixing strategy drastically reduces privacy without any trade-off with utility. 
Indeed, mixing the updates of the model has no impact on the result of the aggregation of the updates computed by the server.
We experimentally evaluate \name and design a new attribute inference attack, \attackss, exploiting the privacy vulnerability of SGD algorithm to quantify privacy leakage in different settings (i.e., the aggregation server can conduct a passive or an active attack).
We show that \name significantly limits the attribute inference compared to a baseline using noisy gradient (well known to damage the utility) while keeping the same level of utility as classic federated learning.
}
\end{abstract}

\newcommand{\name}{\textsc{MixNN}\xspace~}
\newcommand{\namess}{\textsc{MixNN}\xspace}
\newcommand{\attackss}{\textsc{$\nabla$Sim}\xspace}
\newcommand{\attack}{\textsc{$\nabla$Sim}\xspace~}

\maketitle
\pagestyle{plain}
\section{Introduction}

The collection of personal data is a subject firmly grounded in public debates.
The growing awareness of the population on privacy issues led to stronger regulations on data protection (e.g., GDPR, HIPAA) and contributed to the appearance of new services making privacy an incentive vector such as privacy-based search engine (e.g., Duckduckgo, Qwant), web browsing (e.g., Web Proxy, Tor, Brave), or mailing (e.g., Protonmail).
These services rely on infrastructures setup 
by companies, nonprofit organizations promoting privacy, or are fully peer-to-peer involving devices of end-users.

However, personal and private data is still the fuel of all desires. 
In this context, Machine Learning (ML) has emerged as a core technology to analyze and provide learning models from large volumes of data and to perform complex tasks such as classifications, predictions or clustering.
The success of ML has driven different providers to launch Machine Learning as a Service (MLaaS) engines to make ML operation easier for anyone, without the cost and time to build in-house infrastructures.
These new services has led to an ever increasing number of new applications or services relying on ML capabilities in different domains such as computer vision, health analytic and speech recognition to name a few. However, it has been showed that ML models may leak information in the training data~\cite{10.1145/2810103.2813677,song2020systematic,zhao2021infeasibility}. 
The fact that many applications using this technology involve the collection and processing of personal and sensitive data has raised privacy concerns~\cite{decristofaro2020overview}.

Despite being popular, the memorization of training data by a ML model is the source of different privacy violations such as membership, property and attribute inference through passive or active attacks.
Membership inference~\cite{Nasr_2019,jayaraman2020revisiting} refers to the capacity of an adversary to identify if a data point (or the data of an individual) has been used to train the target model. This attack has a serious privacy implication if the model is training with sensitive information (e.g., data from people with certain health status). Property inference~\cite{10.1145/3243734.3243834,zhang2020datasetlevel}, in turn, corresponds to the inference by an adversary of the properties of training data such as the features that characterize each class. This property inference can also concern a subset of the training inputs. 
This ability to learn from training data is desired if the inference is directly related to the main task of the model.
By contrast, attribute inference~\cite{melis2018exploiting} corresponds to the fact that an adversary is able to infer an unintended and undesired attribute not correlated to class's characteristic feature. 
Root causes related to these attack surfaces as well as the link between utility (e.g., through model overfitting~\cite{song2020overlearning,8429311}) and privacy are not well understood.

Recently, Federated Learning (FL)~\cite{bonawitz2019towards} 
has emerged as promising privacy-by-design alternatives to decentralized learning schemes. In such a collaborative scheme, personal data never leaves the user device. Instead, devices (computing and refining a learning model with their own data) and a central server (aggregating models) work together to build a global learning model.
This new ML scheme has attracted many attention these last years, not only from the research community but also from major Internet companies, suggesting future deployments. For instance, Google envisioned to massively exploit FL in the near future (e.g., through its FLoC API~\cite{floc}). 
While the FL scheme is a clear step forward towards enforcing users’ privacy, it still suffers from a large ML-based attack surface including membership, property and attribute inference from participants or from the server.
Different protection mechanisms to limit inference capabilities of an adversary have been proposed~\cite{naseri2021robustness}. 
%
For instance, some solutions~\cite{vandermaaten2020tradeoffs, yu2020salvaging} are based on perturbation in order to reveal only a noisy information to the server, such as differential privacy. However, these solutions significantly damage the accuracy of the model and its capacity to converge. 
Secure aggregation relying on a cryptographic scheme has been also proposed~\cite{bonawitz2016practical,10.1145/3133956.3133982, 10.1145/3372297.3417885}. 
Similar to \namess, this solution ensures that the server is only aware of the aggregate of all models, keeping the model of each participant (and the associated inference) private. 
Beside the overhead of this solution remains low, the underlying cryptographic scheme requires the participation of the server in the protection. 
We argue that such solutions are not deployed in practice.
Indeed, few companies accept to afford the additional cost of the protection. For instance, Private Information Retrieval (PIR) protocols which follow similar cryptographic scheme to protect the profile of users are not widely adopted in practice. 
Moreover, a curious or malicious server trying to infer information from participants will certainly not adopt such a protection.


In this paper, we present \namess, a new privacy-preserving service for FL against inference attacks from a curious or malicious aggregation server.
To achieve that, \name relies on a proxy mixing the layers of the model updates among participants before sending them to the aggregate server.
Like Mixnets to ensure anonymity in information routing~\cite{10.1145/358549.358563}, mixing the layers of the participants' updates of neural network prevents inference attack without decreasing the accuracy of the aggregated model. This solution, albeit simple, leads to drastically improving the privacy without any trade-off with utility. 
In addition, \name is transparent to the FL service, participants only need to configure a web proxy for the associated traffic. To make the deployment of \name easier by anyone (e.g., operate by an individual or non profit organizations willing to protect privacy) and possibly on an untrusted infrastructure, the proxy mixing the neural network layers is running inside an SGX enclave ensuring confidentiality and attestation on its behavior.

To illustrate the capability of \name to protect privacy while maintaining the same level of utility, we design \attack a new attribute inference attack exploiting the privacy vulnerability of the Stochastic Gradient Descent (SGD) algorithm.
More precisely, the server infers sensitive attributes of participants from their model updates.
These updates represent the gradient vectors which minimize their contribution to the model’s training loss.
These gradient vectors are consequently influenced by the local data of the user.
Based on auxiliary information on the model updates from participants belonging to each class of sensitive attributes, \attack uses the gradient vector as a fingerprint to infer attributes.
\attack can be conducted passively only though the observation of model updates (i.e., the case of a curious and undetectable adversary server), or actively by influencing the model sent to participants to extract more information (i.e., the case of a malicious adversary server modifying the protocol).


Our work takes a quantitative and empirical approach.
We implement \name and experimentally evaluate it with several datasets and neural networks architectures.
We also leveraged \attack to quantity privacy leakage through attribute inference attack on classical federated learning scheme, a baseline using perturbation (i.e., noise) to protect the model updates (widely used in Differential Privacy), and \name.
We show that \name limits the attribute inference compared to other baselines 
without decreasing the accuracy of the global aggregated model.
Moreover, we show that the \name proxy introduces only a small latency on the model updates.
The code of \name and \attack are publicly available.


The remainder of this paper is organized as follows.
Section~\ref{back} presents background, Section~\ref{sec-problem} defines the problem and the threat model, Section~\ref{mixnn} explains the design and the implementation of \namess, Section~\ref{sec:attack} presents the new attribute inference attack, Section~\ref{eval} reports the evaluation of \namess, Section~\ref{related} reviews related work, and Section~\ref{ending} concludes this paper.


\section{Background}
\label{back}

In this section, we review background related to Neural Networks~\ref{back-nn}, Federated Learning~\ref{back-fl}, inference attacks~\ref{back-attacks}, Mixnet~\ref{back-mixnet}, and Intel SGX~\ref{back-tee}.

\subsection{Neural Networks}
\label{back-nn}
An ML model is a function $f(\theta): X \mapsto Y$ 
parameterized by a set of parameters $\theta$, where X denotes the input (or feature) space ($X=x_1,x_2,...,x_k$), and Y the output space ($Y=y_1,y_2$).
Training an ML model corresponds to find the optimal set of parameters $\theta$ that fits the training data. This is done by optimizing an objective function (loss) which penalizes the model when it is wrong. For instance, if we consider a classification task trained through a supervised learning, parameters $\theta$ are updated if the model misclassifies training data.

\begin{figure}[!htb]
\centering
\includegraphics[width=0.5\linewidth]{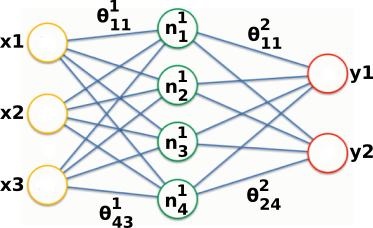}
\caption{Example of Neural Network.} 
\label{fcn}
\end{figure}

Neural networks are a family of ML models which have become popular for a variety of ML tasks. 
A neural network is composed of multiple layers of non-linear mappings from input to intermediate hidden states (or hidden layers) and then to output where each layer transforms the output of the preceding layer to produce input for the next layer.
The topology of the connections between layers and the type of considered transformation function are task-dependent and impact the accuracy of the model. 
For instance, convolutional layers account for locality where each neuron receives a restricted input space 
while a neuron receives the entire previous layer in a fully connected layer.



A neural network $f$ is composed of a collection of $n$ hidden layers ($f=(l_1,l_2,...,l_n)$). Each layer $l_i$ is composed of a set of $m$ neurons ($l_i = (n^i_1, n^i_2, ... n^i_m)$). For input $X$, the output of the neural network,  can be formally written as:
$$f(\theta) = F_n( F_{n-1}( ... F_2(F_1(X)))),$$

where $X$ is the input, $n$ the number of layers, $F_i$ represents a transformation function of the layer $l_i$, and $\theta$ is the set of floating-point weights associated with each connection between two neurons of different layers. 
Considering a fully connected neural network (as depicted Figure~\ref{fcn}), 
$\theta_{ab}^t$ represents the weight connecting the node $n^t_a$ to the node $n^{t-1}_b$.
These weights are updated during training according to the method to optimize the objective function. In this work, we consider Stochastic Gradient Descent (SGD) to do this optimization.
SGD is an iterative approach where the optimizer receives a batch of training data and updates the model parameters $\theta$ at each iteration according to both the direction of the gradient of the objective function and a learning rate $\eta$ which scales the update.
Once the gradient is close to zero, the model has converged to a local minimum and the training is finished. The model is evaluated through its accuracy over testing data points not used to train the model.
The hyperparameters refer to the set of tunable parameters not related to the neural network (e.g., weights associated to connection) such as the number of training iterations, the size of the training batch or the learning rate. 

\subsection{Federated Learning}
\label{back-fl}


Federated Learning (FL) is a collaborative learning scheme to train an ML model~\cite{bonawitz2019towards}. 
In such a scheme, personal data never leaves the device of participants.
Instead, devices train a ML model locally and interact with a central server to build a global learning model.

\begin{figure}[tbp]
	\centering
	\includegraphics[width=\columnwidth]{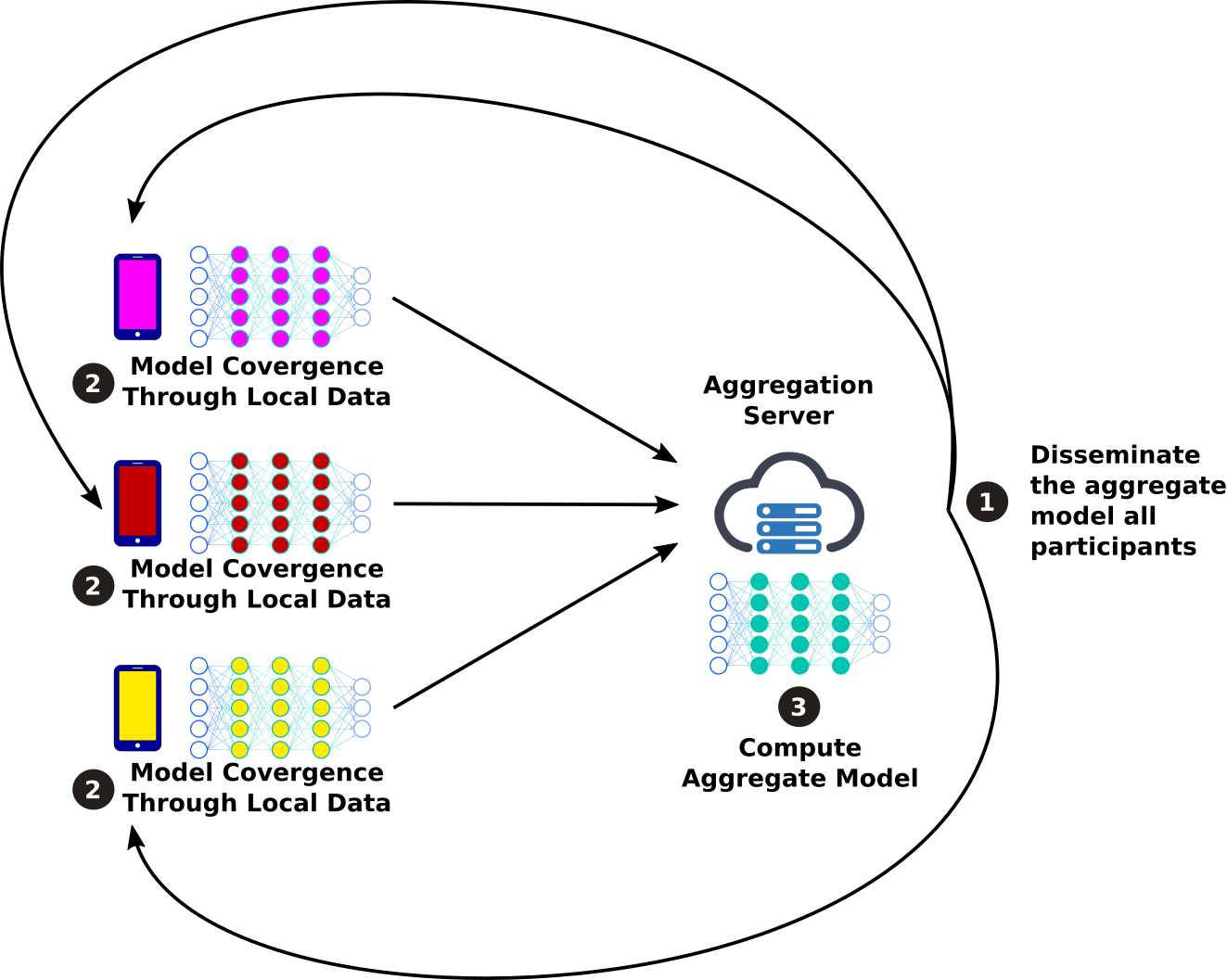}
	\caption{Operating flow of Federated Learning.}
	\label{fig:classic-fl}
	\vspace{-2mm}
\end{figure}

The iterative-based operating flow of classical FL is depicted Figure~\ref{fig:classic-fl}.
Each iteration contains three steps.
First, the aggregation server disseminates a global model to participants (step \ding{182} in the figure).
Each participant then trains and refines this model with its own data stored locally (step \ding{183}).
After this local training, each participant holds its own variation of the model sent by the server.
Participants then send their updated model parameters to the aggregation server.
Finally, the server aggregates all these updates to generate a new global model (step \ding{184}) which will be disseminated to participants in the next iteration.
Iteratively, the global model maintained on the server converges without requiring access to the personal data of participants.

\subsection{Inference Attacks}
\label{back-attacks}


By keeping locally data of the users on their device, FL improves privacy by design.
However, FL can disclose sensitive information via model updates that are based on the training data.
Indeed, any useful ML model reveals something about the population from which the training data was drawn. Indeed, a classifier model for instance may reveal the features that characterize a given class or help construct data points that belong to this class. 
The first privacy violation is property inference: identification of the features that characterize each class, making it possible to construct representatives of these classes through model inversion attacks~\cite{10.1145/2810103.2813677}.
Another privacy violation is attribute inference~\cite{song2020overlearning}: the leak of personal and unintended information (i.e.,  properties that hold for certain subsets of the training data, but not generically for all class members).
The last privacy violation in our setting is membership inference~\cite{shokri2017membership}: given an exact data point, determine if it was used to train the model. 

Memorization of training data by deep neural networks enables an adversary to conduct all these privacy violations. Firstly, this memorization usually combined with over-fitting of the model are exploited by an adversary to conduct a membership inference attack in order to discriminate if a user has been part of the training or not~\cite{Nasr_2019}. This attack has a serious privacy implication especially if the learning model is related to sensitive information (e.g., presence of a certain pathology). Secondly, as deep-learning models come up with separate internal representations of all kinds of features, some of which are unpredictable and independent of the task being learned, the memorization of the training data can be leveraged by an adversary to infer a sensitive attribute~\cite{melis2018exploiting}. In addition, due to the distributed nature of FL, passive and active inference attacks can be conducted by any participant or by the server.

Introducing Differential Privacy in the Stochastic Gradient Descent (DP-SGD)~\cite{Abadi_2016,naseri2021robustness} has been proposed to reduce the inference capability of an adversary, however this solution significantly damages the accuracy of the model and its capacity to converge~\cite{vandermaaten2020tradeoffs, yu2020salvaging}. In addition, the noise calibration and the management of the privacy budget is not trivial.
Other defenses propose to reduce the overfitting~\cite{salem2018mlleaks} but inherently decrease the utility.

\subsection{Mixnets}
\label{back-mixnet}

The concept of mixing information to make them indistinguishable or unlinkable is not new.
Mix networks (Mixnets)~\cite{10.1145/358549.358563} uses this concept to provide a proxy-based anonymity system.
This system aims to provide unlinkability between the message sent by an user, and the message received by the destination.
More precisely, to prevent traffic analysis attacks, Mixnets route each message of the user through a set of anonymity servers called mixes.
Mixes collect and shuffle (or mix) many messages before to route then to the destination.
A variety of mixnets have been proposed including Aqua~\cite{10.1145/2534169.2486002}, Riffle~\cite{DBLP:journals/popets/KwonLDF16}, and Mixminion~\cite{1199323} addressing differently the tradeoff between anonymity, latency and bandwidth.

The limitation of these systems are similar to Tor, it is difficult for a user to determine which edge is uncompromised and powerful adversaries controlling both ends of the circuit can still deanonymize clients.

\begin{figure*}[tbp]
	\centering
	\includegraphics[width=1.5\columnwidth]{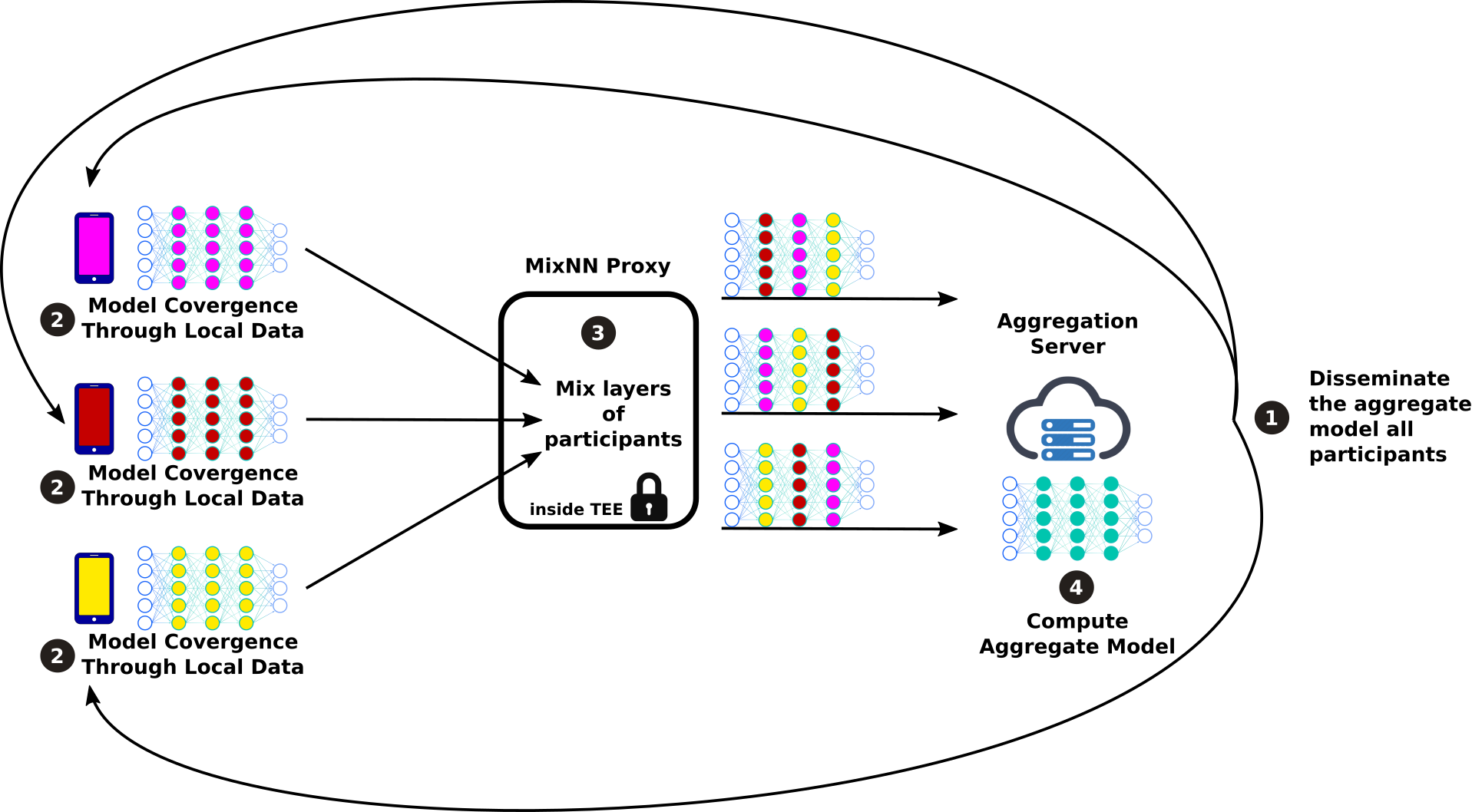}
	\caption{\name introduces a proxy which receives the parameter updates from each participant, shuffle them to remove attribute footprint before to route them to the aggregation server.}
	\label{fig:mixnn-fl}
\end{figure*}

\subsection{Intel SGX}
\label{back-tee}

The \name proxy relies on a Trusted Execution Environment (TEE), which leverages custom microprocessor zones, to enforce isolation, confidentiality and integrity of code and data.
Specifically, we use Intel Software Guard Extensions (SGX)~\cite{costan2016intel,anati2013innovative} which defines the concept of enclave.
The memory of an enclave is encrypted and cannot be directly accessed by other system software even by privileged code (e.g., the operating system or hypervisor).
Enclaves can be attested to prove that the code running in the enclave is the one intended, and that it is running on a genuine Intel SGX platform. Once attested, enclaves can be provisioned with secret data by using authenticated secure channels. Moreover, enclaves can persist secret data outside the trusted zone by using a sealing mechanism.
However, such protection comes with resource constraints. 
More precisely, only 96 MB out of the 128 reserved for the enclave can be used by applications. Although virtual and dynamic memory support is available~\cite{chen2017presage,chen2016princess,mckeen2016intel}, it incurs significant overheads in paging (i.e., the sealing and unsealing operations used an encryption key derived from the CPU hardware).

\section{System and adversary model}
\label{sec-problem}

Before presenting \name and our attribute inference attack \attackss, 
we describe our assumptions and the considered threat model.
The operating flow of \name involves three premises with different level of trust, namely: (i) the client machine; (ii) the \name proxy; and (iii) the aggregation server.

First, we assume that the client machine is trusted. 
This includes the training data and all the computations performed locally.
We do not consider malicious users trying to poison the model or to introduce backdoors.

Second, we assume that the \name proxy is running inside an Intel SGX enclave on an untrusted node.
An adversary is thus not able to compromise the behavior or the data of the proxy but can monitor the node (i.e., honest but curious), possibly physically (e.g., monitoring network traffic, power consumption or memory access patterns).
Consequently, an adversary can leverage side channel attacks~\cite{206170} to infer information.
We assume that the SGX enclave has generated a public and private key pair ($k_{pub}$ and $k_{priv}$).

Lastly, we consider a malicious aggregation server.
This server builds a model for a main classification task through a federated learning scheme but also aims to infer sensitive attributes from participants.
This aggregation server can either conduct a passive or an active attack.
Specifically, it can passively follow the FL operational flow to infer sensitive attributes or abuse the protocol by sending a specific model to each participant to amplify the possible inference.
We do not consider any mechanism on participants to detect misbehavior of the server.
In addition, we consider protected exchanges between participants and the \name proxy.

We consider 
an adversary able to collect or to use a public dataset with similar raw data (including the sensitive attribute) in order to build attack models.
Each of these attack models is trained only with data from one specific class of sensitive attribute (e.g., one attack model for activity detection trained only with data from men, and another one trained only through data from women).
These specialized attack models allow the adversary to compare for each participant the parameter models (i.e., the direction of the gradient obtained from the local training) to the direction of the gradient which would result in attack models. 
The attribute inference is computed based on a similarity metric between these directions as detailed Section~\ref{sec:attack}.


\section{\name Framework}
\label{mixnn}

In this section, we first present an overview of the \name (Section~\ref{sec:mixnn-overview}), the equivalence in terms of utility with a classical FL (Section~\ref{sec:mixnn-equivalence}), and then present implementation details (Section~\ref{sec:mixnn-implem}).

\subsection{Overview}
\label{sec:mixnn-overview}

To avoid inference attacks during the learning process of a service using a federated learning scheme, \name operates as depicted in Figure~\ref{fig:mixnn-fl}.
To use \namess, users have only to configure its system to use a proxy for the associated traffic (e.g., through the configuration of its browser). 
As such, users seamlessly get protected without changing their habits. 
More precisely, compared to the classical FL pipeline (Figure~\ref{fig:classic-fl}), all parameter updates will be sent to the \name proxy instead of the aggregation server.
To secure these updates, they are encrypted with the public key of the enclave (i.e., $k_{pub}$) to ensure that only the \name proxy is able to read and process them.
Once loaded in the enclave, the proxy decrypts and stores the parameter updates of each layer in different lists.
The proxy then pick at random one update for each layer in the associated list to generate the message containing the parameter updates to send to the aggregation server.
Note that the proxy needs to initialize first each list with $k$ updates before to send updates to the aggregation server.

The rest of the workflow remains unchanged compared to the classical one.
The server aggregates the parameter updates to generate a global model which will be disseminated to all participants.
Participants will then refine this model locally with their personal data before sending the parameter updates to the \name proxy.

The accuracy of the global model remains unchanged with or without using \name.
Indeed, whether mixed or not, the aggregation of parameter updates of each layer are identical.
In contrast, the privacy leakage through the footprint of parameter updates returned by participants is drastically  reduced.
Specifically, by receiving an update mixing information from different users (breaking potential attribute footprints), the aggregation server is not able to infer any sensitive attribute.
Consequently, \name is able to drastically improve privacy without compromising the accuracy of the system (i.e., no trade-off between utility and privacy).

\subsection{Utility Equivalence}
\label{sec:mixnn-equivalence}
By design, \name provides the same utility than a classical FL scheme. In this section, we prove this equivalence.

Let $C$ be the number of participants sending their updates to the proxy. We show in this section that whether the participants use \name or not, the resulting aggregated model is the same. We assume that the considered \name proxy has enough information to send $L$ updates to the server. Then the proxy creates a sequence $\left(M_{ij}\right)$ such that $\forall (j_1,j_2) \in \{1, \cdots,n\}^2 \text{ with }j_1\neq j_2$ and $\forall i \in \{1, \cdots, L\} \quad M_{ij_1} \neq M_{ij_2}$.
And also such that $\forall (i_1, i_2) \in \{1, \cdots, L\}^2 \text{ with } i_1 \neq i_2$ and $\forall j \in \{1, \cdots, n\} \quad M_{i_1j} \neq M_{i_2j}$.

According to previously defined notation for the nodes of a neural network, we define the $t$-th layer of the $c$-th participant of the proxy by $\left(\theta_..^t\right)^c$. In the following matrix, each line is a model sent by the proxy. We remark that each combination of participant/layer appears once and only once in the matrix. This is a fundamental assumption regarding the equality of accuracy level between traditional FL and \name.

\begin{equation*}
\footnotesize
A = 
\left(
    \begin{matrix}
    \left(\theta_..^1\right)^{M_{11}} & \left(\theta_..^2\right)^{M_{12}} & \cdots&\left(\theta_..^n\right)^{M_{1n}}\\
    \left(\theta_..^1\right)^{M_{21}} & \left(\theta_..^2\right)^{M_{22}} & \cdots&\left(\theta_..^n\right)^{M_{2n}}\\
    \vdots &\vdots&\ddots&\vdots\\
    \left(\theta_..^1\right)^{M_{L1}} & \left(\theta_..^2\right)^{M_{L2}} &\cdots &\left(\theta_..^n\right)^{M_{Ln}}\\
    \end{matrix}
    \right)
\end{equation*}

Now, with the regular FL procedure, the information sent by the participant is:

\begin{equation*}
\footnotesize
B = 
\left(
    \begin{matrix}
    \left(\theta_..^1\right)^{1} & \left(\theta_..^2\right)^1 & \cdots&\left(\theta_..^n\right)^1\\
    \left(\theta_..^1\right)^2 & \left(\theta_..^2\right)^2 & \cdots&\left(\theta_..^n\right)^2\\
    \vdots &\vdots&\ddots&\vdots\\
    \left(\theta_..^1\right)^C & \left(\theta_..^2\right)^C &\cdots &\left(\theta_..^n\right)^C\\
    \end{matrix}
    \right)
\end{equation*}

We note $Agr : \mathcal{M}(C\times n) \longrightarrow \mathcal{M}(1\times n)$ the aggregation function which makes the mean of the columns.  We show that $Agr(A) = Agr(B)$
\begin{equation*}
\footnotesize
Agr(A) = 
    \left(
    \frac{1}{L}\sum_{i=1}^L \left(\theta_..^1\right)^{M_{i1}},  \frac{1}{L}\sum_{i=1}^L \left(\theta_..^2\right)^{M_{i2}}, \cdots,  \frac{1}{L}\sum_{i=1}^L\left(\theta_..^n\right)^{M_{in}}
    \right)
\end{equation*}

\begin{equation*}
\footnotesize
    Agr(B) = 
    \left(
    \frac{1}{C}\sum_{c=1}^C \left(\theta_..^1\right)^{c},  \frac{1}{C}\left(\theta_..^2\right)^c, \cdots,  \frac{1}{C}\sum_{c=1}^C\left(\theta_..^n\right)^c
    \right)
\end{equation*}

We make the additional assumption that $L=C$ which means that the \name proxy waits for the $C$ participants to send their updates before mixing. Which gives us that 
\begin{equation*}
\footnotesize
Agr(A) = Agr(B) 
    \iff 
\end{equation*}

\begin{equation*}
\footnotesize
\left[\forall l \in \{1,\cdots,L\} \quad \sum_{c=1}^C\left(\theta_..^l\right)^{M_{cl}} = \sum_{c=1}^C\left(\theta_..^l\right)^c\right]
\end{equation*}

Which is true since our assumption on $\left(M_{ij}\right)$ gives us that $\varphi : \{1,\cdots,C\} \longrightarrow \{1,\cdots,C\} \quad c \mapsto M_{cl}$ is a bijective mapping.

\subsection{Implementation}
\label{sec:mixnn-implem}

\name is implemented inside an Intel SGX enclave to protect its behaviors and confidentially even if it is deployed on an untrusted node.
In addition, the parameter updates are encrypted by participants with the public key $pub$ of the enclave.
Each update received by the \name proxy, once decrypted, is split by layer and the parameters associated to each layer are stored in different lists.
The size of these lists (noted $k$) and the memory allocation according to the considered neural network models are initialized at the creation of the enclave.
The $k$ first parameter updates are used to fill out the different lists.
Once these lists are full, for each further received parameter update, the \name proxy picks at random and removes one element in each list to build a parameter update to send to the aggregation server. The empty element in each list is then filled out with information coming from the incoming update.

To avoid side-channel attacks against SGX~\cite{206170}, 
the cost (i.e., the execution time) to process an update is constantly the same.
Depending on the considered model, the size of a model can be important and not fit into the memory limit of the enclave (96MB), requiring encrypted storage outside the enclave.
To avoid side-channel attack based on memory access, ORAM mechanisms (e.g., 
ZeroTrace~\cite{DBLP:conf/ndss/SasyGF18}) can be adopted to carry out secure and oblivious access of data. The associated overhead is negligible in our context where updates are sent only periodically.

\section{\attackss: A Similarity-Based Attribute Inference Attack}
\label{sec:attack}

\begin{figure}[tb]
\centering
{
  \includegraphics[width=0.45\textwidth]{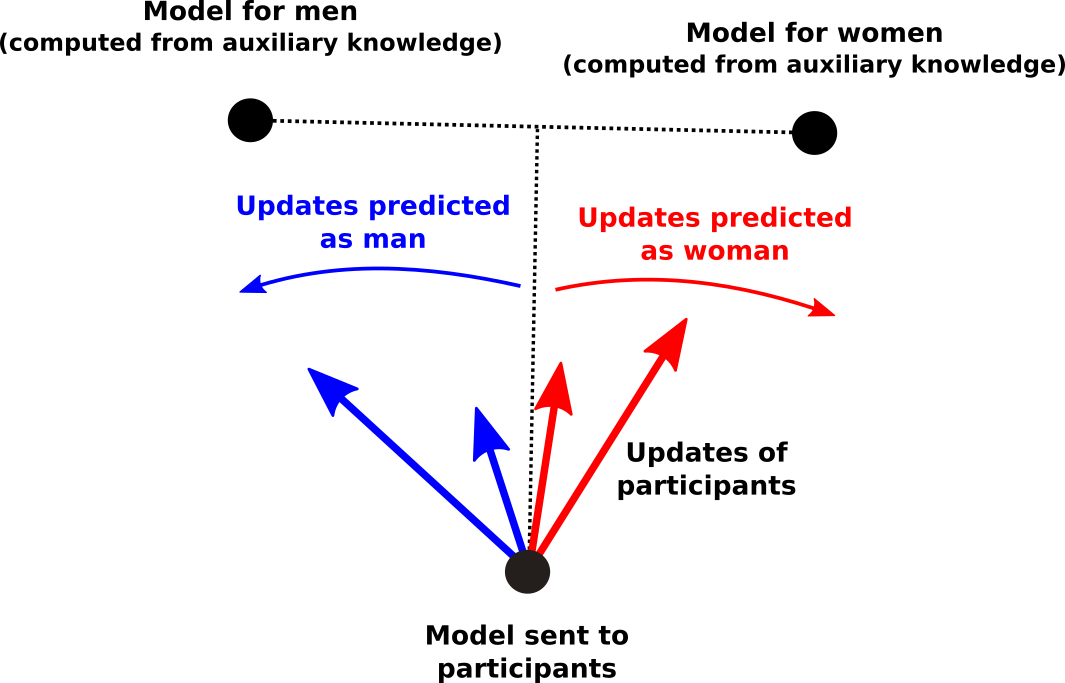}}
\caption{\attack~  infers attributes according to the gradient vector returned by participants (i.e., the parameter updates) and the learning models representative to each class of sensitive attributes (background knowledge).}
\label{fig:attack}
\end{figure}

We design a new attribute inference attack, \attackss, exploiting the privacy vulnerability of the SGD algorithm.
Specifically, \attack is based on a similarity metric measuring the gradient vector returned by participants to minimize its contribution to the model’s training loss (i.e., the parameter update).
Figure~\ref{fig:attack} illustrates how the inference attack works.
During one round, the gradient vector returned by a participant reflects how its local data have influenced the global model disseminated by the aggregation server at the beginning of the round.
This gradient vector is used as a fingerprint to infer the sensitive attribute.
This fingerprint can be amplified if the attack is conducted during multiple rounds.
Indeed, in this case, the influence of the user data on the evolution of the global model is more observable.

More formally, \attack measures the cosine similarity between the gradient vector returned by a participant and a reference gradient vector representative to a specific sensitive class of attribute (e.g., representative to men or women).
This reference gradient vector is computed through the background knowledge of the adversary. As described in Section~\ref{sec-problem}, we consider an adversary able to learn a classification model (similar to the one used for the main task) trained only with participants with a specific sensitive attribute.
\attack evaluates for which of the representative models of each class of sensitive attribute the returned gradient vector is closest.

\attack can be active or passive. In its passive form, the curious aggregation server follows the standard FL learning round and compares the model update returned by participants to the models representative to the sensitive attributes computed from auxiliary knowledge. In contrast, in its active form, the aggregation server sends to participants the model calculated for being equidistant from the models associated to the sensitive attributes (e.g., the model for men and women). In this case, the server is malicious and actively modifies the protocol to conduct the inference attack.

\section{Evaluation}
\label{eval}

We now report the results in terms of utility (Section~\ref{eval-utility}) and privacy (Section~\ref{eval-privacy}) provided by \name under the considered experimental setup (Section~\ref{eval-setup}).
We also analyse the robustness of \name against an aggregation server trying to defeat the protection. (Section~\ref{eval-robustness}) as well as its performance from a systems perspective (Section~\ref{eval-sysperf}).

Our results show that \name efficiently reduces the information leakage through an attribute inference attack without compromise on the accuracy of the model. 
We also show that \name introduces a negligible end-to-end latency.


\subsection{Experiment Setup}
\label{eval-setup}

In this section we presents the experimental setup used to
evaluate \namess, which includes datasets, metrics,  the comparative baseline we compared against, and the considered methodology.


\subsubsection{Dataset} We used two image recognition benchmark datasets (CIFAR10 and LFW) and two motion datasets for activity recognition (MotionSense and MobiAct) to assess \namess.

 %

\textbf{CIFAR10} is a major image classification benchmarking dataset where the data records are composed of 60,000 32$\times$32 RGB images where each record is mapped to one of 10 classes of common objects such as airplane, bird, cat, dog. There are 50,000 training images and 10,000 test images.
The main task is the classification of the images.
We artificially define 20 participants split into three groups with different preferences. We define 3 types of preference which corresponds to specific and non overlapping categories of images.
The dataset is slightly balanced, two groups gather 6 participants and the last one gathers 8 participants.
The profile of the participant is composed of 80\% of images corresponding to its preferred classes, and the remaining 20\% is composed of random images from other classes.
The sensitive attribute is the preferences of the user.
%
%
%

\textbf{MotionSense}~\cite{ortiz2015smartphone} contains data captured from an accelerometer (i.e., acceleration and gravity) and gyroscope at a constant frequency of 50Hz collected with an iPhone 6s kept in the front pocket. Overall, a total of 24 participants have performed six activities (i.e., going downstairs, going upstairs, walking, jogging, sitting and standing) during 15 trials in the same environment and conditions.
The main classification task is the activity detection and the sensitive attribute is the gender of the users.

\textbf{MobiAct}~\cite{mobiact} records the motion data from 58 subjects during more than 2500 trials, all captured with a Samsung Galaxy S3 in the front pocket. This dataset includes signals recorded from the accelerometer and gyroscope at 20Hz. We only used the trials corresponding to the same activities as MotionSense in order to do the evaluation with the same settings.
Similar to MotionSense dataset, the main classification task is the activity detection and the sensitive attribute is the gender of the users.


\textbf{Labeled Faces in the Wild (LFW)}~\cite{LFWTech} contains face images for face recognition with 13,233 total samples with images for 5,749 people.
The dataset additionally has attributes such as age, race, gender, smile, facial hair, glasses etc.
The main classification task is smile detection and the sensitive attribute is the gender of the users.

For CIFAR10, MotionSense and MobiAct datasets, we use a neural network composed of two convolutional layers and three fully connected layers for the classification task. For LFW, in turn, we use a more complex architecture provided by Facebook, named Deep Face~\cite{taigman2014deepface}. This neural network is composed of multiple convolutional, locally connected, maxpooling, and fully connected layers.




\begin{figure*}[tb]
\minipage{0.5\textwidth}
\centering
\subfloat[CIFAR10]{\label{fig:accuracyCIFAR10}{
  \includegraphics[width=1\textwidth]{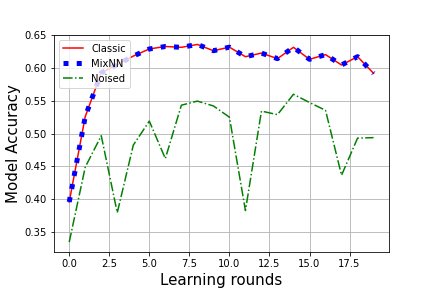}}}
  
\vspace{-2mm}
\subfloat[MotionSense]{\label{fig:accuracyMS}{
  \includegraphics[width=1\textwidth]{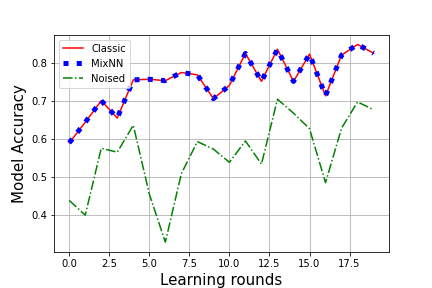}}}

\endminipage\hfill
\minipage{0.5\textwidth}
\vspace{-2mm}
\subfloat[MobiAct]{\label{fig:accuracyMS}{
  \includegraphics[width=1\textwidth]{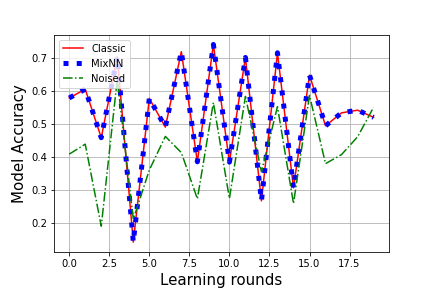}}}

\vspace{-2mm}
\subfloat[LFW]{\label{fig:accuracyMS}{
  \includegraphics[width=1\textwidth]{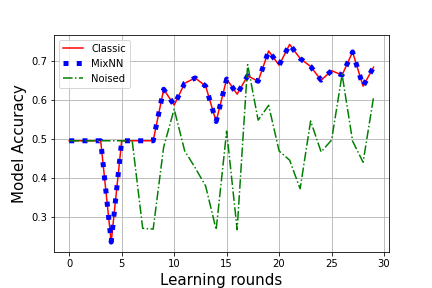}}}


\endminipage
\caption{\name provides the same utility than a standard  FL scheme, noisy gradient however decreases significantly the utility and slows down the convergence.}
\label{fig:utility}

\end{figure*}




\subsubsection{Evaluation Metrics}


We evaluate \name through three complementary dimensions: utility, privacy and system performance.

To evaluate the utility of the target model, we consider the classification accuracy for the main task (e.g., the activity detection), noted $Model$ $Accuracy$, measuring the ratio of number of correct predictions to the total number of predictions made. 

\begin{figure*}[tb]
\centering
\minipage{0.5\textwidth}
\centering
\subfloat[CIFAR10]{\label{fig:accuracyCIFAR10}{
  \includegraphics[width=0.9\textwidth]{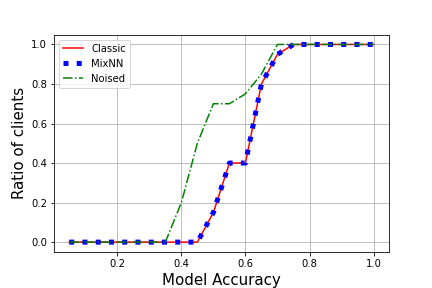}}}
  
\subfloat[MotionSense]{\label{fig:accuracyMS}{
  \includegraphics[width=0.9\textwidth]{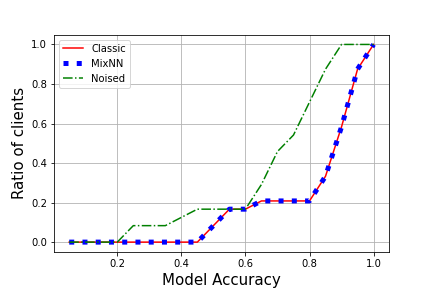}}}

\endminipage\hfill
\minipage{0.5\textwidth}

\subfloat[MobiAct]{\label{fig:accuracyMS}{
  \includegraphics[width=0.9\textwidth]{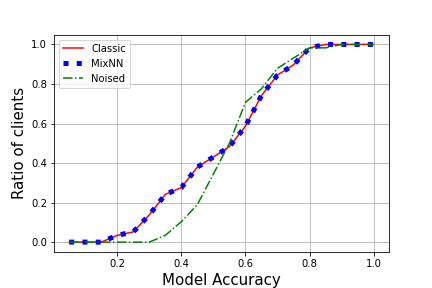}}}
  
\subfloat[LFW]{\label{fig:accuracyMS}{
  \includegraphics[width=0.9\textwidth]{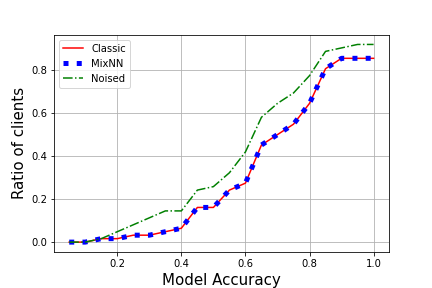}}}
\endminipage
\centering
\caption{Using noisy gradient decreases the utility for all participants.}
\label{fig:utilitycummulative}

\end{figure*}


We use \attack to conduct the inference attack and we use the classification accuracy of the sensitive attribute 
to estimate the success of the attribute inference, noted $Inference$ $Accuracy$.
The value of this metric indicates a data leakage according to the number of classes and if the dataset is balanced over all classes.
For instance, with a balanced dataset over the gender, an accuracy above 50\% indicates a data leakage through attribute inference attack. This indicates that the adversary is able to identify the gender of a participant with an accuracy higher than random guess.








To evaluate the behavior of \name from a systems perspective, we consider the end-to-end latency which is the time spent by the proxy to route the parameter updates to the aggregation server. 

\subsubsection{Baselines}
We compare the utility and privacy provided by \name against a comparative approach using noisy gradient widely used in Differential Privacy studies~\cite{naseri2021robustness,zhu2019deep}.
We consider an implementation based on an introduction of Gaussian noise to the updates computed through a classical local training such as using in local differential privacy~\cite{naseri2021robustness}.


\subsubsection{Methodology}

The dataset is split between training and testing, with 5/6 of trials used for training and validation and 1/6 for testing. For CIFAR10, the federated learning model is trained on 3 local epochs for a size of data batch of 32 samples on each learning rounds, the server aggregates 16 users on each of the 10 learning rounds. 
For MotionSense (and MobiAct), the training is (respectively) done on 2 (and 3) local epochs for batches of 256 (and 64) samples for each of the 20 learning rounds, and the server aggregates 20 users for MotionSense and 40 users for MobiAct.
For LFW, the training is done on 2 local epochs for batches of 16 samples for each of the 30 learning rounds, and the server aggregates 20 users. 
For every datasets, we use the "Adam" optimizer proposed by Tensorflow.
We use 5-fold cross-validation in which the testing set is randomly generated from 1/5 of the users. Reported results correspond to average over 5 repetitions of each experiment.
The experiments have been computed on a Laptop DELL, intel core i7 (i7-6600U) and 4G of RAM, using TensorFlow version 2.4.
The noise introduced consists on adding a Gaussian noise $\mathcal{N}(0,1)$ on each scalars of the neural network weights.
The attack models are trained for 5 learning rounds of the previous architecture and use 4/5 users as background knowledge. 

\subsection{No compromise with utility}
\label{eval-utility}

In this section, we evaluate the capacity of \name to protect privacy without compromising the utility.
We compare the accuracy performance for the main classification task provided by \name against a classic FL scheme (i.e., without \name proxy) and a baseline using noisy gradient such as using in local differential privacy.
Figure~\ref{fig:utility} reports the accuracy according to the learning round for all datasets.
 First, the results show that the same level of accuracy is provided by a standard FL scheme and \name with an accuracy growing according to the learning rounds.
This result is expected due to the aggregation equivalence of both approaches.
Second, the results show that noisy gradient provides 10\% lower accuracy on average and slows down the convergence.
Figure~\ref{fig:utilitycummulative} reports the cumulative distribution of the accuracy over the population of participants at the learning round 6.
Results show that most of the participants have an accuracy with noisy gradient smaller than \name for all datasets (on average 0.56 for noisy gradient against 0.68 for \namess).

\begin{figure*}[tb]

\minipage{0.5\textwidth}
\centering
\subfloat[CIFAR10]{\label{fig:accuracyCIFAR10}{
  \includegraphics[width=0.9\textwidth]{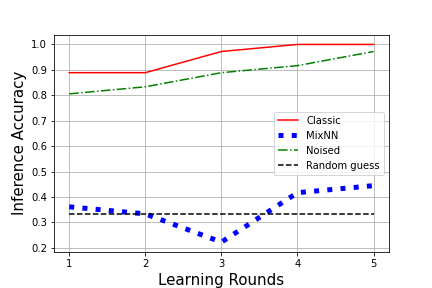}}}
  
\subfloat[MotionSense]{\label{fig:accuracyMS}{
  \includegraphics[width=1\textwidth]{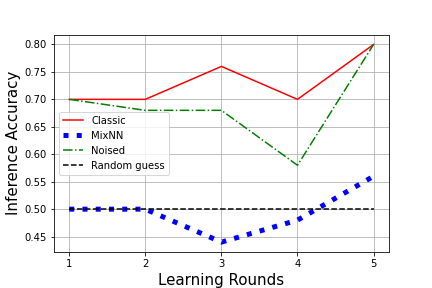}}}

\endminipage\hfill
\minipage{0.5\textwidth}

\subfloat[MobiAct]{\label{fig:accuracyMS}{
  \includegraphics[width=0.9\textwidth]{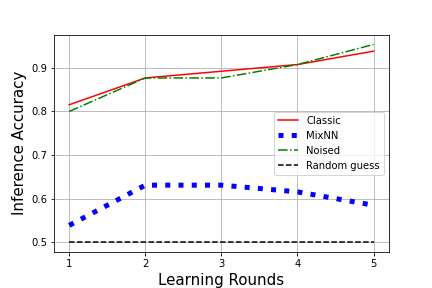}}}
  
\subfloat[LFW]{\label{fig:accuracyMS}{
  \includegraphics[width=0.9\textwidth]{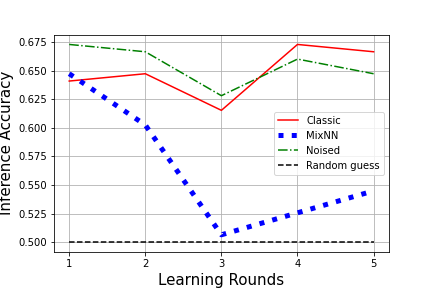}}}
\endminipage
\caption{\name better prevents attribute leakage compared to using noisy gradient.}
\label{fig:privacy}

\end{figure*}

\begin{figure*}[tb]

\minipage{0.5\textwidth}%
\centering
\subfloat[CIFAR10]{\label{fig:accuracyCIFAR10}{
  \includegraphics[width=0.9\textwidth]{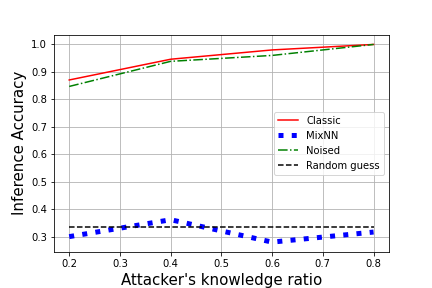}}}
  
\vspace{-2mm}
\subfloat[MotionSense]{\label{fig:accuracyMS}{
  \includegraphics[width=0.9\textwidth]{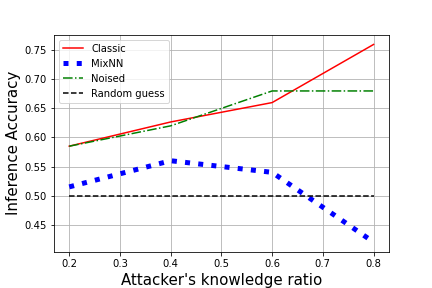}}}

\endminipage\hfill
\minipage{0.5\textwidth}

\subfloat[MobiAct]{\label{fig:accuracyMS}{
  \includegraphics[width=0.9\textwidth]{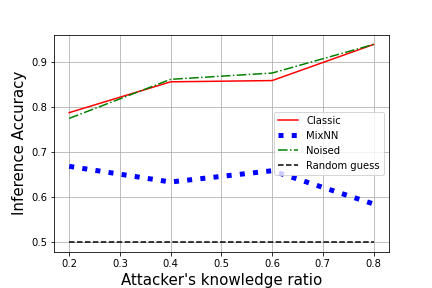}}}
  
\subfloat[LFW]{\label{fig:accuracyMS}{
  \includegraphics[width=0.9\textwidth]{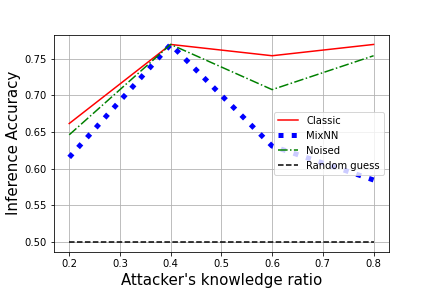}}}
\endminipage
\caption{With more background knowledge, the malicious aggregation server infers more information with a standard FL scheme or using noisy gradient. This background knowledge has only a small impact on the protection of \namess.}
\label{fig:privacyRatio}

\end{figure*}

\subsection{Prevent information leakage}
\label{eval-privacy}

In this section, we evaluate the privacy leakage through the sensitive attribute attack \attack through an active attack (i.e., which represents the worst case where the aggregation server is malicious and sends a calibrated model to participants to amplify the inference).
Figure~\ref{fig:privacy} reports for all datasets the accuracy of the inference for \namess, Classical FL and noisy gradient according to a growing number of learning rounds.
First, results show that without any protection, the server can infer the sensitive attribute of participant with a quasi perfect accuracy (100\% of accuracy after 4 rounds for CIFAR10, and around 80\%, 94\%, and 66\% of accuracy after 5 rounds for MotionSense, MobiAct, and LFW dataset, respectively.
This means that the gradient vector returned by participants and exploited by \attack provides an efficient footprint to infer attributes.
Second, results show that \name successfully limits the privacy leakage with an inference accuracy close to a random guess (0.33 for CIFAR10, and around 0.5 for MotionSense, MobiAct and LFW).
The precision of the inference tends to increase with the number of local learning round but becomes stable once the model is converged.
Finally, results show that noisy gradient leaks less information than a standard FL scheme but much more information than \name for all datasets
(on average around 65\% more inference accuracy).

As described Section~\ref{sec:attack}, \attack leverages background knowledge to build a model representative to men and a model representative to women. The attack then measures the distance between these representative models and the model updates returned by participants, the closest distance indicating a predominance towards a sensitive attribute.
We evaluate now the impact of this background knowledge on the accuracy of the inference.
Figure~\ref{fig:privacyRatio} depicts the inference accuracy according to a growing ratio of data used as background knowledge to build the representatives models of the sensitive attributes (the quality of a model is usually correlated to the volume of data used in learning, the larger, the better). As expected, a model built with more background knowledge is more representative to individuals with a specific sensitive attribute and consequently improves the accuracy of the inference for both a classic FL and a solution using noisy gradient.
However, results show that \name successfully protects participants against inference attack regardless the quantity of background knowledge used by the aggregation server.

\subsection{Robustness of the protection}
\label{eval-robustness}

\begin{figure}[tbp]
	\centering
	\includegraphics[width=0.9\columnwidth]{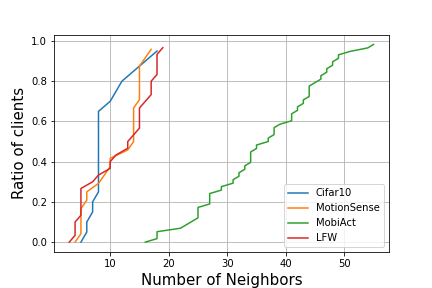}
	\caption{Many participants have very close model updates making it difficult for a malicious aggregation server to retrieve and distinguish all pieces of the gradient coming from the same participant once mixed by \namess.}
	\label{fig:robustness-radius}
\end{figure}

\name shuffles model updates sent by participants.
A malicious aggregation server could then try to break the protection by enumerating through possible combinations of the shuffled updates to "reconstruct" back the original update for instance.
Figure~\ref{fig:robustness-radius} reports for all datasets the cumulative distribution over all participants of the number of neighbors who have a gradient very close (i.e., in a radius of 0.5 using euclidean distance). All participants have at least a few other alter egos with very close gradients making it difficult for a malicious aggregation server to retrieve and distinguish all pieces of the gradient (i.e., the layers of the neural network) coming from the same participant once mixed by \namess.

\subsection{System performance}
\label{eval-sysperf}



We implemented a \name proxy inside an SGX enclave to evaluate its system performance.
To do that, the enclave receives the update parameters in an encrypted binary format and then decrypts and stores each layer of the updates in the trusted memory of the enclave.
%
We ran the experiment using the model designed for CIFAR10 (i.e., two convolutional layers and three fully connected layers). Each update consumes 26.9MB inside the enclave and is processed in 0.19s (0.17s for the decryption and 0.02s for the storage). The mixing operation spends only 0.03s on average in our experiment.
This processing time and the memory consumption are dependent on the size of the model. Using a model with three convolutional layers and three fully connected layers slightly increases this time to 0.22s and 51.3MB. 
However, the constant processing time over all updates for a given model (i.e., the \name proxy waits to receive a constant number of model updates before to mix them) reduces the surface for side channel attacks.
As the learning round of classical FL scheme is usually not conducted at high rate (e.g., only when the device is plugged in and has a WiFi connection), this short delay introduced by \name is negligible.

\section{Related work}
\label{related}


The incentive behind using FL is to collectively build a learning model with better accuracy than if each user trained a model with their own data. The goal is to improve the accuracy as much as possible but several dimensions have an influence. The standard FL scheme~\cite{mcmahan2017communicationefficient} learns one global model and replicates it locally on every client. However heterogeneity of data across user devices can severely degrade performance of standard federated averaging for ML learning applications, especially for atypical users. Indeed, one unique model cannot cope with the heterogeneity of data and provide the best utility for all users~\cite{dysan}. To address this data heterogeneity, several approaches have been proposed such as local adaptation~\cite{yu2020salvaging, arivazhagan2019federated}  and clustering~\cite{sattler2019clustered}. 
%
Specifically, the clustering mechanism proposed in~\cite{sattler2019clustered} also leverages a similarity metrics between the model updates sent back by participants (similar to \namess) to cluster the population.


\cite{Nasr_2019} designs passive and active inference algorithms for federated learning. However, this work only targets membership inference. In addition, while this attack also exploits the privacy vulnerabilities of the SGD algorithm, authors used a neural network to classify if a participant is a member of the training data or not a member.
\cite{zhu2019deep} also exploits the gradient exchange to infer private training data of participants. 
To do that, authors iteratively optimize "dummy" inputs and labels to minimize the distance between dummy gradients and real gradients. Once the optimization finished, the dummy data is close to the private training data.
\cite{jagielski2020auditing}, in turn, investigates the guarantees of differentially private SGD but via data poisoning attacks.

Running \name in an Intel SGX enclave improves trust and confidentiality through an isolated execution environment.
However, this TEE is still vulnerable to side channel attacks~\cite{206170, sgaxe}.
The most common countermeasure is to use data oblivious algorithms.
The objective of this technique is to eliminate the link between the nature of data inputs and the execution of the program (e.g., through the execution time or memory footprints). 
To achieve that, the obfuscation technique consists to hide potential patterns by making them all uniform regardless of the considered data. 
To reduce its inherent cost, the considered data oblivious algorithm needs to be chosen carefully according to the application~\cite{alam2020sgxmr}. 

\section{Conclusion}
\label{ending}

We presented \namess, a proxy-based privacy-preserving framework to prevent attribute inference attacks conducted from a curious or malicious aggregation server exploiting the model updates.
\name breaks the attribute footprint leaked in the model updates by mixing layers between multiple participants.
As this mixing strategy does not impact the result of the model aggregation performed by the server, the privacy improvement of \name does not compromise the utility of the model learned collaboratively.
We also designed \attackss, a new attribute inference attack which exploits the privacy vulnerability of the SGD algorithm. This attack can be passive or active. 
We experimentally evaluated \name with a standard image recognition bench-mark dataset and compared it against a state-of-the-art baseline using local differential privacy.
Results show \name provides the same model accuracy than a classical FL scheme (i.e., the same utility) while providing a better protection against attribute inference attack (i.e., a better privacy). 


\bibliographystyle{unsrt}
\bibliography{sample-base}

\end{document}